# Light-YOLOv8-Flame: A Lightweight High-Performance Flame Detection Algorithm

Jiawei Lan, Ye Tao*, Zhibiao Wang, Haoyang Yu, and Wenhua Cui

*Abstract*—Fire detection algorithms, particularly those based on computer vision, encounter significant challenges such as high computational costs and delayed response times, which hinder their application in real-time systems. To address these limitations, this paper introduces Light-YOLOv8-Flame, a lightweight flame detection algorithm specifically designed for fast and efficient real-time deployment. The proposed model enhances the YOLOv8 architecture through the substitution of the original C2f module with the FasterNet Block module. This new block combines Partial Convolution (PConv) and Convolution (Conv) layers, reducing both computational complexity and model size. A dataset comprising 7,431 images, representing both flame and non-flame scenarios, was collected and augmented for training purposes. Experimental findings indicate that the modified YOLOv8 model achieves a 0.78% gain in mean average precision (mAP) and a 2.05% boost in recall, while reducing the parameter count by 25.34%, with only a marginal decrease in precision by 0.82%. These findings highlight that Light-YOLOv8-Flame offers enhanced detection performance and speed, making it well-suited for real-time fire detection on resource-constrained devices.

*Index Terms*—Flame Detection; YOLOv8 Optimization; Lightweight Model; FasterNet Block

## I. INTRODUCTION

IN recent years, with the rapid development of industrialization and urbanization, public safety concerns have become increasingly prominent, particularly with the growing frequency and severity of fires and their associated damages. Traditional fire detection methods, while capable of providing a certain level of early warning, exhibit significant limitations in terms of response speed, false alarm rate control, and accuracy in fire source localization, especially in complex environments. Consequently, enhancing the accuracy and operational efficiency of flame detection algorithms, especially for implementation on embedded platforms, has emerged as a topic of significant engineering relevance.

At present, flame image detection approaches are generally classified into two primary types: conventional methods and those based on deep learning. Conventional approaches primarily rely on digital image processing techniques, focusing on analyzing the color, shape, and dynamic characteristics of flames. Researchers have employed various color spaces, such as RGB, HIS, YUV, and Lab, to extract flame color features and apply specific thresholds to distinguish flame regions, thus enabling flame recognition. In 2010, Gu Junjun et al. [1] identified flames by analyzing multiple features, including flame area, circularity, and the number of sharp angles. In 2013, Lai Xiaojun et al. [2] used cameras equipped with visible light and infrared filters to capture infrared images. By combining circularity and sharp angle variations of flames, they achieved rapid and accurate fire detection. Similarly, Liang-Hua Chen et al. [3] introduced a vision-driven algorithm that utilizes color, spatial layout, and temporal dynamics to detect fire areas in video sequences. The method uses a Gaussian mixture model to detect fire-colored pixels and employs spatio-temporal features to eliminate spurious regions, achieving robust fire detection across varying conditions. In 2019, Marcia Baptista et al. [4] introduced the CICLOPE system, a tele-surveillance platform for real-time smoke and fire detection. In 2020, Khalil A. et al. [5] developed a novel method combining RGB and Lab color models to enhance fire detection accuracy through motion detection and flame object tracking. Although these traditional methods perform well in simple environments, they are prone to false positives and missed detections in complex scenarios, such as varying lighting conditions and smoke interference. Furthermore, these techniques often rely on manual feature extraction and threshold setting, limiting their generalizability.

In contrast, deep learning-based methods offer significant advantages by automatically extracting flame features and accurately identifying and localizing flames in complex environments. These methods are particularly valuable for real-time performance and accuracy in fire detection, making them highly applicable for fire prevention and early response systems. In 2015, Polednik et al. [6] employed Deep Convolutional Neural Networks (CNNs) with the Caffe framework to detect fires in images and videos. In 2017, Huttner et al. [7] proposed a deep learning-based fire detection system using Google's Inception V3. Their work evaluated various optimizers, loss functions, learning rates,

This work was supported by the National Natural Science Foundation of China (62272093), the Economic and Social Development Research Topics of Liaoning Province (2025-10146-244), the Postgraduate Education and Teaching Reform Research Project of Liaoning Province (LNYJG2024092), and the Undergraduate Innovation Training Program of University of Science and Technology Liaoning.

Jiawei Lan is an undergraduate student of School of Computer Science and Software Engineering, University of Science and Technology Liaoning, Anshan, China. (e-mail: 3203035719@qq.com).
Ye Tao is an Associate Professor of School of Computer Science and Software Engineering, University of Science and Technology Liaoning, Anshan, China. (Corresponding author to provide phone: +86-133-0422-4928; e-mail: taibeijack@163.com).
Zhibiao Wang is an undergraduate student of School of Computer Science and Software Engineering, University of Science and Technology Liaoning, Anshan, China. (e-mail: wangzhibiao24@mails.ucas.edu.cn).
Haoyang Yu is an undergraduate student of School of Computer Science and Software Engineering, University of Science and Technology Liaoning, Anshan, China. (e-mail: 3223428129@qq.com).
Wenhua Cui is a Professor of School of Computer Science and Software Engineering, University of Science and Technology Liaoning, Anshan, China. (e-mail: taibeijack@126.com).

and convergence times to optimize performance. In 2019, Aslan et al. [8] proposed a real-time fire recognition approach utilizing Deep Convolutional Generative Adversarial Networks (DCGAN). By adopting a sequential training scheme and incorporating temporal-spatial flame variation features, they achieved flame detection in videos with exceptionally low false alarm rates. In 2022, Ding Hao et al. [9] developed an enhanced flame detection model based on YOLOv3, aiming to improve the extraction of dynamic shape features. Their method integrates ResNet50_vd as the backbone network, along with Deformable Convolutional Modules and Intersection over Union (IoU) aware modules to improve flame feature extraction. Similarly, Wang Yuanbin et al. [10] developed a fire detection method using a CNN optimized with dropout inactivation probabilities, addressing poor generalization and low detection accuracy by predicting optimal inactivation probabilities for different convolution layers. In 2023, Sun Xiaoqing et al. [11] proposed a fire detection algorithm built on the YOLOv4 framework, emphasizing real-time detection and accurate flame recognition. This methodology was specifically designed for efficient fire warning systems, demonstrating substantial advancements in both detection speed and precision. In 2024, Xie Kangkang et al. [12] improved flame detection accuracy under the YOLOX framework by incorporating the Swin-T backbone network, a weighted Bidirectional Feature Pyramid Network (BiFPN), and the Coordinate Attention (CA) mechanism. Although deep learning techniques have significantly advanced the field of flame detection, mitigating many of the limitations of traditional methods in complex environments, several challenges remain. These include achieving high accuracy in backgrounds with diverse environmental conditions, improving sensitivity to small and distant flames, which are often difficult to detect due to their lower intensity and smaller size, and balancing real-time performance with resource consumption, particularly in resource-constrained settings.

Despite the progress made in modern fire detection technologies compared to traditional methods, flame detection algorithms still face issues such as delayed response times, high false alarm rates, and challenges in accurately locating the fire source, limiting their effectiveness in practical applications. This study proposes Light-YOLOv8-Flame, an optimized YOLOv8 model in which the original C2f module is replaced with a FasterNet Block module. This block combines Partial Convolution (PConv) and Convolution (Conv) layers, resulting in improvements to mean average precision (mAP) and recall rate while achieving a lightweight model that enhances speed and efficiency, making it suitable for deployment on edge devices.

## II. FLAME DETECTION DATASET

The flame image detection dataset utilized in this study consists of 7,431 images, which include scenes of indoor fires, wildfires, vehicle fires, as well as non-flame images from indoor environments, forests, vehicles, streetlights, and sunsets. Figure 1 illustrates the process of constructing the flame image detection dataset.

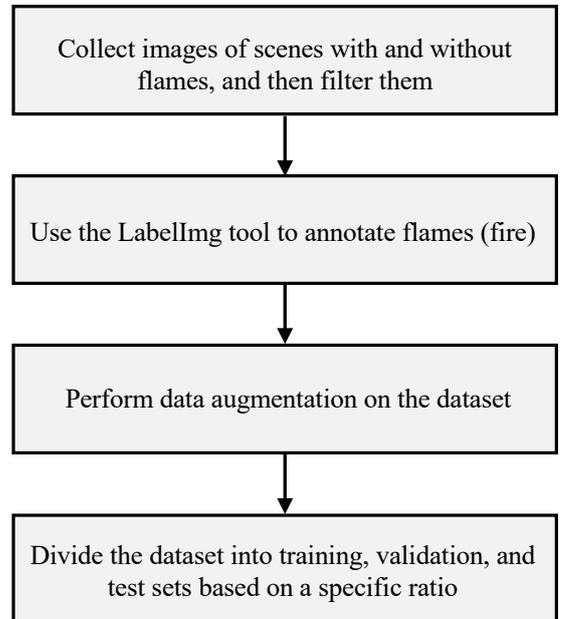

Fig. 1. Flame image dataset creation process

### A. Data Collection

This dataset is essential to deep learning, as it facilitates both the training and testing phases and serves as the foundation for evaluating algorithm performance. To address challenges in existing fire datasets, such as limited scene diversity, interference from smoke or similar objects, and difficulties in detecting small targets, this study collected approximately 3,000 images from platforms such as CSDN and GitHub. These images encompass both flame and non-flame scenes, ensuring a varied and comprehensive dataset. After organizing and filtering the data, 2,477 high-quality images containing flames were selected. Using data augmentation techniques, a final experimental dataset comprising 7,431 images was constructed. The dataset's image samples are visually displayed in Figure 2.

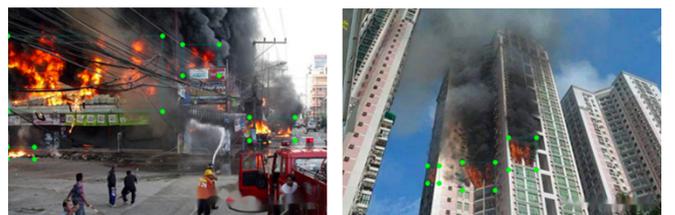

Fig. 2. Partial flame image detection dataset

### B. Dataset Annotation

The labeling guidelines for this dataset require flames to be annotated as comprehensively as possible. In cases where a flame is partially obscured by a small object or if two flames overlap, they are considered a single flame. The dataset follows the YOLO format, with all images stored in *.jpg format. The original images were annotated using the labelImg tool, as shown in Figure 3. Users can load the folder containing the images to be labeled via the tool's "Open" option, while the "Open Dir" function provides access to the folder where annotation files are stored, saved in *.txt format. The labeling process begins by selecting "Create RectBox," and the results are automatically saved upon completion of the annotations.

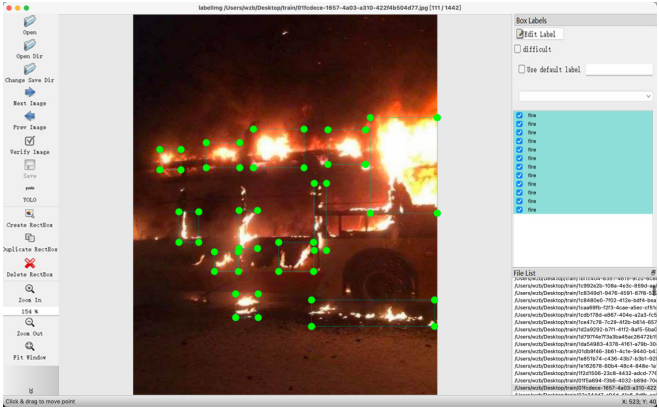

Fig. 3.　LabelImg annotation tool interface

Upon completing the annotation of all images, corresponding label files are generated, which contain the positional information for each target in the original images. As shown in Figure 4, the positional information for each target is recorded on a separate line in the *.txt file. If multiple targets are present, their information is sequentially listed within the same *.txt file. Each line contains the target's class label (x_class), the center position of the bounding box represented by (x_center, y_center), and its spatial dimensions—width and height—all expressed in normalized coordinates. The first number, 0, denotes the label for flames.

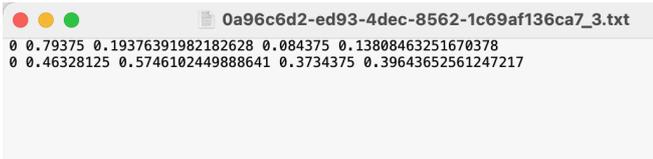

Fig. 4.　Example of a label file in .txt format

### C.　Data Augmentation

Data augmentation is a crucial technique for expanding a dataset by modifying existing data or generating new data based on the original dataset. This approach addresses issues of data scarcity, enhances the model's adaptability to diverse scenarios, provides key invariant features, and reduces the likelihood of overfitting, thereby enhancing its capacity to generalize across varied inputs. Common geometric transformations include image flipping, rotation, cropping, scaling, translation, and jittering, which simulate variations in object positioning, size, and orientation. Pixel-level transformations, such as the addition of salt-and-pepper noise, Gaussian noise, Gaussian blur, adjustments in the HSV color space, changes in brightness and contrast, histogram equalization, and white balance adjustments, help the model handle lighting variations and distortions. The data augmentation techniques employed in this study primarily include the following:

*Spatial Transformation*

Since neural networks interpret different representations of the same object as new samples, this study simulates object appearances at varying sizes and angles by employing horizontal flipping, mirroring, and random cropping on the images. These transformations enrich the dataset and enhance the algorithm's generalization ability. The resulting effects are illustrated in Figure 5.

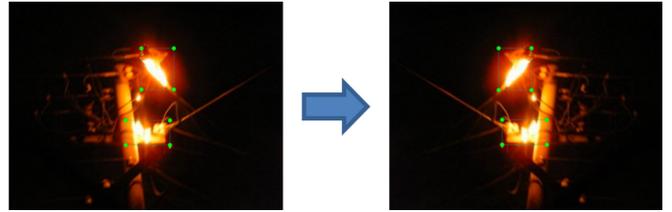

Fig. 5.　Result of spatial transformation

*Random Occlusion*

Random occlusion is used as a regularization technique to prevent overfitting by substituting specific areas of an image with random pixel values or the average pixel values of the training set. The resulting effects are demonstrated in Figure 6.

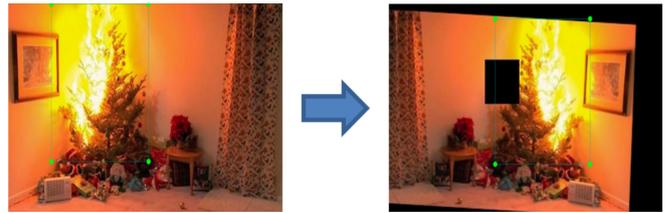

Fig. 6.　Result of random occlusion

*Adding Noise*

The addition of noise enhances data diversity by introducing random perturbations into the original images. This technique simulates various imaging conditions and environmental interferences, enhancing the model's capacity to generalize and maintain robustness. The resulting effects are shown in Figure 7.

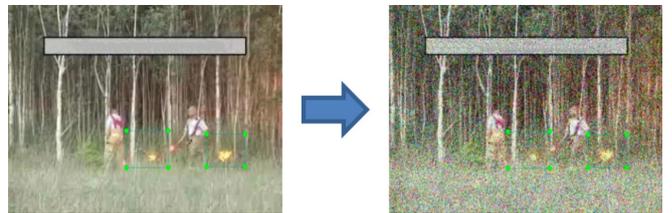

Fig. 7.　Result of noise addition

*Adjusting Brightness*

By adjusting the brightness of the images, this technique simulates diverse lighting conditions, improving the model's adaptability to different environmental contexts. The resulting effects are illustrated in Figure 8.

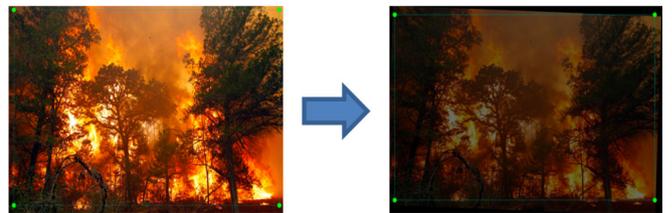

Fig. 8.　Result of brightness adjustment

By applying augmentation strategies including rotation, partial obstruction, and cropping, the number of specific samples is increased. High-quality images are then selected for classification and statistical analysis. If a particular scene contains too few negative samples, the number of negative samples in that scene is augmented. Similarly, if a scene has an insufficient number of positive samples, the number of positive samples is increased accordingly.

*D. Dataset Splitting*

Following data augmentation, the training set was increased to 6,192 images. The dataset was then partitioned into training, validation, and test subsets at a 10:1:1 ratio, as detailed in Table I.

TABLE I
DATASET PARTITIONING

| Category | Number of Images (Count) |
|---|---|
| Training Set | 6196(1549*4) |
| Validation Set | 617 |
| Test Set | 617 |
| Total | 7431 |

To organize the dataset, create a folder named fires_dataset_enhancement within the ./cfg/dataset directory of YOLOv8, following the directory structure depicted in Figure 9. The annotated *.jpg and *.txt files generated by labelImg should be organized into the respective training, validation, and test sets, placed in the train, val, and test directories. Additionally, ensure that the dataset path is correctly specified in the *.yaml file.

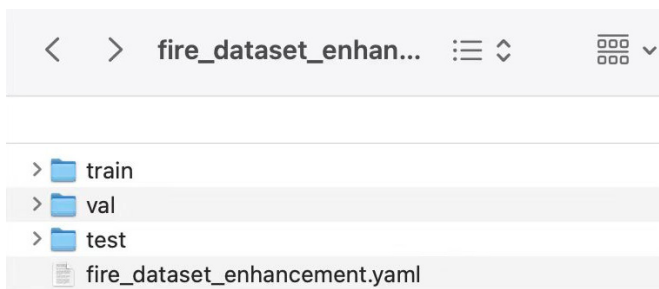

Fig. 9. Directory structure of the dataset

### III. Algorithm Design

*A. YOLO Object Detection Algorithm*

YOLO (You Only Look Once) [13] is a widely recognized one-stage object detector that formulates object localization and classification as a unified regression task, generating bounding boxes and category outputs directly from input images. Since its initial release, the algorithm has experienced several version upgrades, each significantly improving its performance and accuracy.

YOLOv1 initially proposed simplifying object detection by converting it into a single regression problem. This effect was realized through image grid division and direct prediction of bounding boxes and category labels. However, it performed poorly in detecting small objects and handling dense scenes. YOLOv2 [14] introduced batch normalization and a high-resolution classifier, and improved detection performance through the anchor box mechanism. YOLOv3 [15] further enhanced the network structure by adopting a multi-scale feature pyramid, improving performance, especially for small object detection. YOLOv4 [16] built upon YOLOv3 by incorporating CSPDarknet53 as the backbone network and integrating advanced activation functions and network structures, which significantly improved both detection accuracy and speed. YOLOv5 further optimized the training and inference processes, making the model more efficient and user-friendly.

YOLOv6 [17] made fine-tuned adjustments to the model's depth and width, enabling it to capture intricate details in complex scenes while maintaining high efficiency. YOLOv7 [18] employed advanced mechanisms and efficient fusion designs to improve its capability in processing complex scenes and large-scale images.

In February 2023, Ultralytics introduced YOLOv8, a multi-task learning framework capable of handling object detection, instance segmentation, and image classification tasks. YOLOv8 offers five versions—YOLOv8n, YOLOv8s, YOLOv8m, YOLOv8l, and YOLOv8x—each designed with different model sizes and computational requirements. While these versions share the same network structure, they vary in depth and width. Due to its lightweight design and low computational demand, while still meeting the accuracy standards for flame detection, YOLOv8s is adopted as the target model.

The backbone network of YOLOv8 is based on CSPDarknet53 and utilizes Depthwise Separable Convolution (DSC) [19] and Residual Blocks (RB) to enhance network efficiency and accuracy. The backbone is divided into two parts: preprocessing and feature extraction. The preprocessing part handles initial image processing tasks, such as scaling and normalization. During the feature extraction stage, structures like depthwise separable convolutions and residual blocks are combined with convolutional layers, batch normalization, and the SiLU activation function to deepen feature extraction. The SiLU activation function, defined in Equation (1), effectively addresses issues of saturated outputs and vanishing gradients. The function's curve is shown in Figure 10.

$$SiLU(x) = \frac{x}{1+e^{(-x)}} \qquad (1)$$

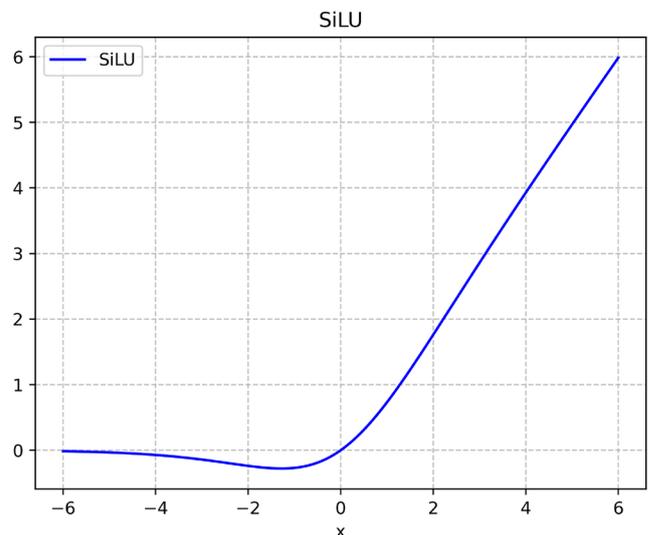

Fig. 10. SILU activation function curve

The specific structure of the C2f module, depicted on the right side of Figure 11, consists of multiple convolutional units and residual structures. This structure improves YOLOv8's efficiency in maintaining gradient propagation, without compromising its lightweight architectural characteristics. Additionally, the SPPF module, also shown on the right side, integrates convolutional and pooling layers, facilitating high-level image feature extraction and improving the network's efficiency and accuracy.

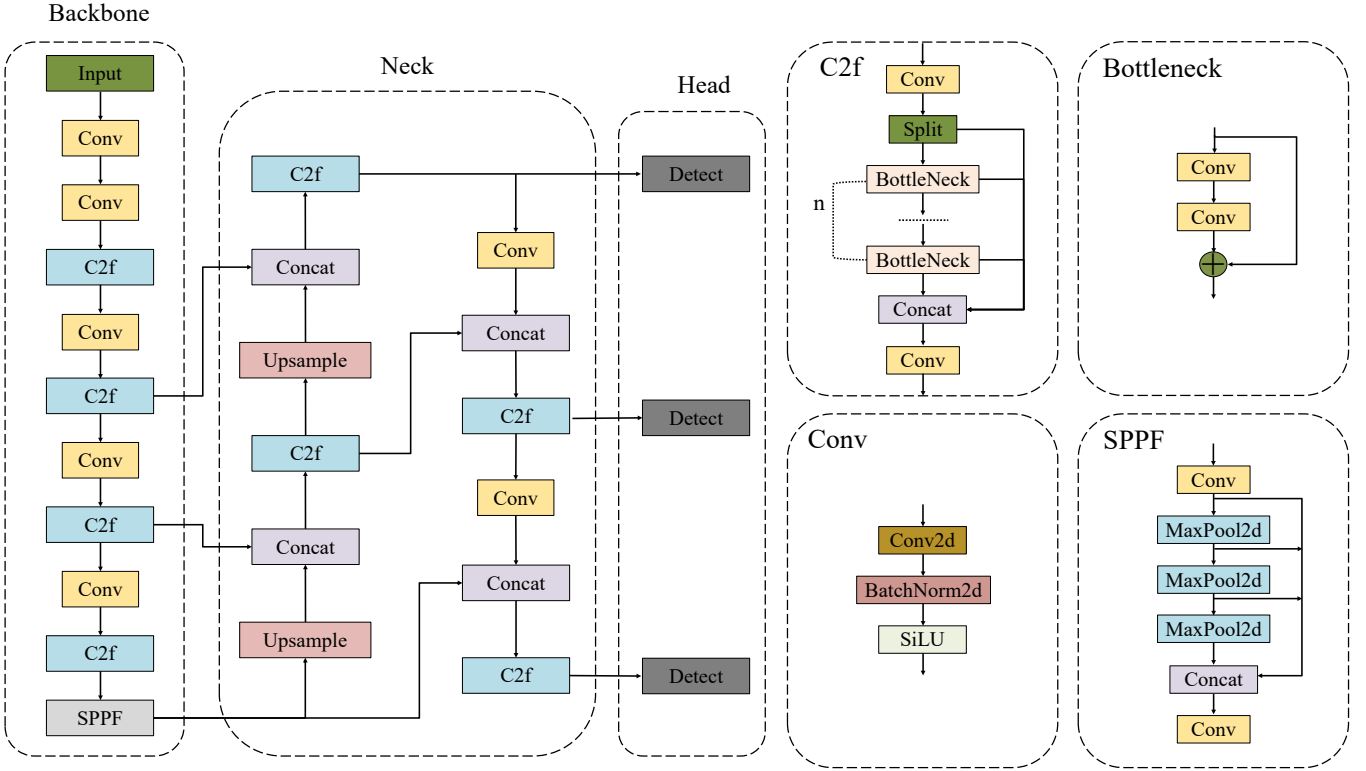

Fig. 11. Architecture of the YOLOv8 network

The neck network is composed of convolutional modules, C2f modules, and upsampling layers, which effectively integrate multi-level feature information to enhance the detection capability for objects of different sizes. The C2f module is employed in both feature fusion and feature extraction processes. The neck network combines the advantages of the Path Aggregation Network (PANet) [20] and Feature Pyramid Networks (FPN) [21]. FPN is responsible for extracting and constructing multi-scale feature maps from deep convolutional layers. By performing top-down upsampling and lateral connections, it preserves object information while leveraging low-level background information. PAN, on the other hand, enhances feature extraction and fusion in a bottom-up manner, further strengthening the model's spatial feature analysis capability.

The detection head uses the SPPF structure, an optimized version of Spatial Pyramid Pooling (SPP), which retains the multi-scale feature extraction capability while significantly enhancing inference speed and efficiency by streamlining the pooling computation process. The SPPF design reduces the computational burden of pooling steps, enabling rapid feature extraction and improving the model's performance in handling objects of various scales. This design is essential in real-time detection contexts, enabling YOLOv8 to efficiently and accurately process high-throughput visual data streams.

### B. Light-YOLOv8-Flame
*FasterNet model*

In lightweight applications for object detection, commonly used models include MobileNet [22], ShuffleNet [23], and GhostNet [24], which leverage Depthwise Convolution (DWConv) [25] or Group Convolution (GConv) [26] to capture spatial features while reducing computational complexity. Although DWConv performs well in reducing model complexity and computational resource requirements, it may not be suitable for all tasks and image resolutions. Therefore, when utilizing DWConv, it is crucial to balance model complexity with performance.

While aiming to reduce GFLOPs, it is also necessary to address the issue of increased memory access. Additionally, the network often includes extra data processing steps, such as concatenation, refinement, and pooling, which are critical for enhancing the performance of lightweight models.

To overcome these limitations, Jierun Chen et al. [27] proposed the FasterNet model. The design goal of this model series is to improve running speed across various devices while ensuring the accuracy of visual recognition tasks remains unaffected. FasterNet improves spatial feature extraction efficiency through minimizing redundant computation and memory operations. As shown in Figure 12, the FasterNet architecture is organized into four hierarchical feature processing stages. Each stage begins with either an embedding layer (using a 4x4 convolution with a stride of 4) or a merging layer (using a 2x2 convolution with a stride of 2) to reduce spatial dimensions and increase feature channels. The PConv module in FasterNet is specifically designed to optimize resource usage, particularly on high-performance computing devices such as GPUs.

PConv is a spatial feature extraction technique in FasterNet that enhances the performance of CNN models while significantly reducing computational redundancy and memory access. As shown in the PConv structure on the left side of Figure 12, it performs conventional convolution operations only on a subset of the input channels, ensuring that the input and output feature maps have the same number of channels while maintaining the method's generality. By applying convolution to only a portion of the input channels, PConv reduces the computational complexity of CNN models while preserving spatial information. This leads to faster training and inference times, as well as improved accuracy in tasks such as image recognition and segmentation. The computational complexity of PConv is given in Equation (2), where $h$ is the channel height, $w$ is the channel width, $c_p$ is the number of continuous network channels, and $k$ is the convolution kernel size.

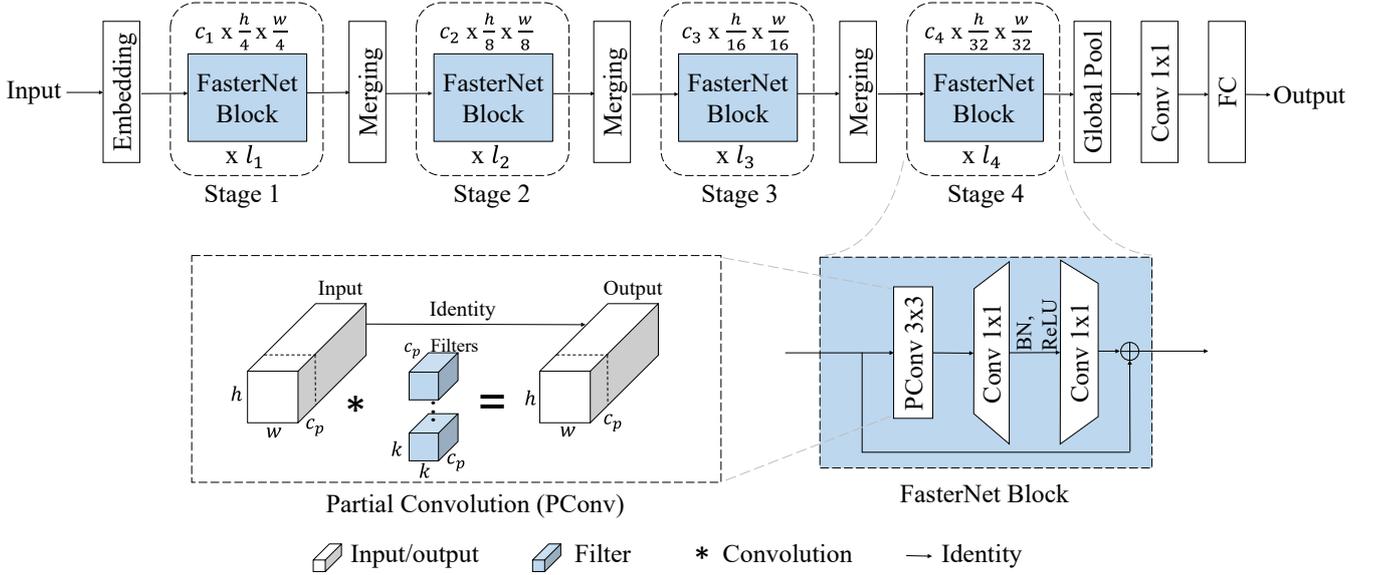

Fig. 12. Architecture of the FasterNet model

$$h \times w \times k^2 \times c_p^2 \qquad (2)$$

If $r = c_p/c$ is set to 1/4, the computational complexity of PConv is only 1/16 that of a full convolution layer. Moreover, PConv requires less memory access, with its complexity expressed in Equation (3):

$$h \times w \times 2c_p + k^2 \times c_p^2 \approx h \times w \times 2c_p \qquad (3)$$

To maximize the utilization of all channel information and enhance feature extraction capabilities, researchers integrated Pointwise Convolution (PWConv) with PConv. This combination forms a T-shaped convolution structure that improves the effective receptive field of the input feature map, focusing more on the central region compared to standard convolution, thereby enhancing the processing of central information. This design not only improves precision in feature extraction but also boosts the model's sensitivity to critical features.

PWConv employs a 1×1 filter size, performing convolution on each pixel of the input using a single scalar value to fully utilize information from all channels. Adding PWConv to PConv significantly enhances performance, especially in tasks involving large-scale image processing. The combined T-shaped convolution emphasizes the center position of the input feature map, as opposed to conventional convolution, which processes features uniformly. PWConv also reduces the dimensionality of the input feature maps, lowering the computational and memory requirements of subsequent convolution layers while improving output quality. The computational complexity of the combined T-shaped convolution is given in Equation (4).

$$h \times w \times \left(k^2 \times c_p \times c + c \times (c - c_p)\right) \qquad (4)$$

The complexity of T-shaped convolution is greater than that of a single PConv or PWConv, as shown in Equation (5), where $(k^2 - 1)c > k^2 c_p$. For instance, when $c_p = \frac{c}{4}$ and $k = 3$, the complexity increases accordingly. However, this structure can be easily implemented in two steps using conventional convolution.

$$h \times w \times \left(k^2 \times c_p^2 + c^2\right) \qquad (5)$$

The FasterNet architecture is divided into four hierarchical stages, with each stage comprising a PConv layer and two PWConv layers, as illustrated in Figure 12. These layers together form an inverted residual structure, where the intermediate PWConv layer has more channels and uses a shortcut connection to reuse the input features.

To maintain feature diversity and reduce processing latency, normalization and activation operations are applied only after the intermediate PWConv layer in each stage. Batch normalization is prioritized in these stages, as it can be combined with adjacent convolution layers, accelerating inference speed while being as effective as other normalization methods. In FasterNet variants, Gaussian Error Linear Units (GELU) [28] are selected as the activation function for smaller models, while Rectified Linear Units (ReLU) [29] are used for larger models.

Additionally, the FasterNet architecture includes embedding layers (using standard 4×4 convolution kernels with a stride of 4) or merging layers (using standard 2×2 convolution kernels with a stride of 2). These layers are responsible for spatial downsampling and the expansion of feature channels. This design allocates more computational load to the last two stages of the model, reducing memory access while increasing FLOPS. Finally, the module integrates a global average pooling layer, a 1×1 convolution layer, and a fully connected layer to complete feature transformation and classification.

The primary design objective of the FasterNet model is to enhance efficiency across diverse devices while maintaining high accuracy. Through careful optimization of the structural design, FasterNet achieves efficient utilization of computational resources, significantly accelerating processing speed without compromising detection accuracy.

When evaluating FasterNet's performance, it was compared with several popular lightweight networks, such as ShuffleNetV2, GhostNet, and MobileNetV2 [30]. Under the same Top-1 ACC metric, FasterNet demonstrated faster processing speeds and shorter computation times, offering a clear advantage in terms of efficiency and responsiveness compared to these models.

*C2f module*

The C2f module is a residual block designed to enhance feature extraction capabilities, drawing inspiration from the C3 module and incorporating principles from the Efficient Long-Range Attention Network (ELAN) [31]. The primary objective of the C2f module is to improve gradient flow capture while maintaining the model's lightweight nature. By utilizing residual connections, the C2f module allows the network to learn input-output relationships more effectively, thus improving the accuracy of feature-level representations. Additionally, due to its relatively simple structure, the C2f module effectively lowers both computational burden and architectural complexity, thereby enhancing the model's runtime efficiency, particularly in resource-constrained environments. Figure 13 provides a depiction of the C2f module's structure.

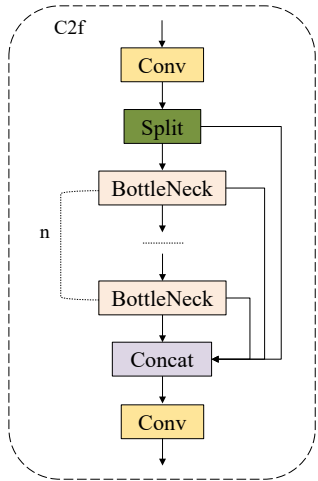

Fig. 13. Structure of the C2f module

As illustrated, the C2f module consists of two convolutional layers connected in series, with a residual connection between them. The residual connection allows input data to be directly passed to the output, while the convolutional layers learn the relationship between the input and output. This design helps mitigate common challenges in deep neural network training, such as vanishing gradients and insufficient representational capacity.

The C2f module used in YOLOv8 includes two types of BottleNeck structures: BottleNeck1, which is used in the Backbone, and BottleNeck2, which is used in the Neck, as shown in Figure 14.

By comparing these two structures, it is evident that BottleNeck1 improves upon the DenseNet architecture by introducing additional cross-layer connections, eliminating convolution steps in the branches, and incorporating a Split operation. This design not only enriches feature information but also reduces the computational load while maintaining performance, thereby achieving a balance between efficiency and effectiveness.

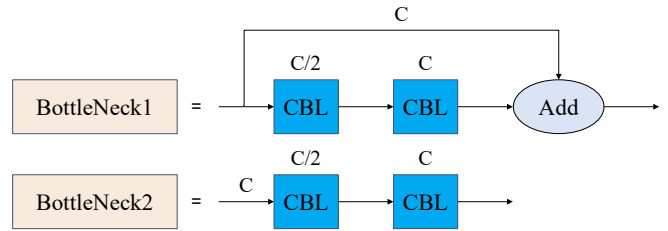

Fig. 14. Two types of bottleneck structures

*Implementation of the Faster-C2f Lightweight Module*

The FasterNet Block integrates PWConv with PConv, enabling efficient utilization of information from all channels, facilitating more diverse feature extraction, and enhancing overall model performance. Figure 15 illustrates a structure composed of one PConv layer and two successive PWConv layers, forming an inverted residual block with an intermediate stage that increases the number of channels.

In the original YOLOv8 backbone network, the C2f module consists of standard convolutional layers and multiple BottleNeck blocks, which include extensive skip connections and additional Split operations. While this design improves performance, it also makes the network structure more complex and computationally intensive, requiring significant computational resources and time. This is not ideal for flame detection tasks, which demand faster processing and lower latency.

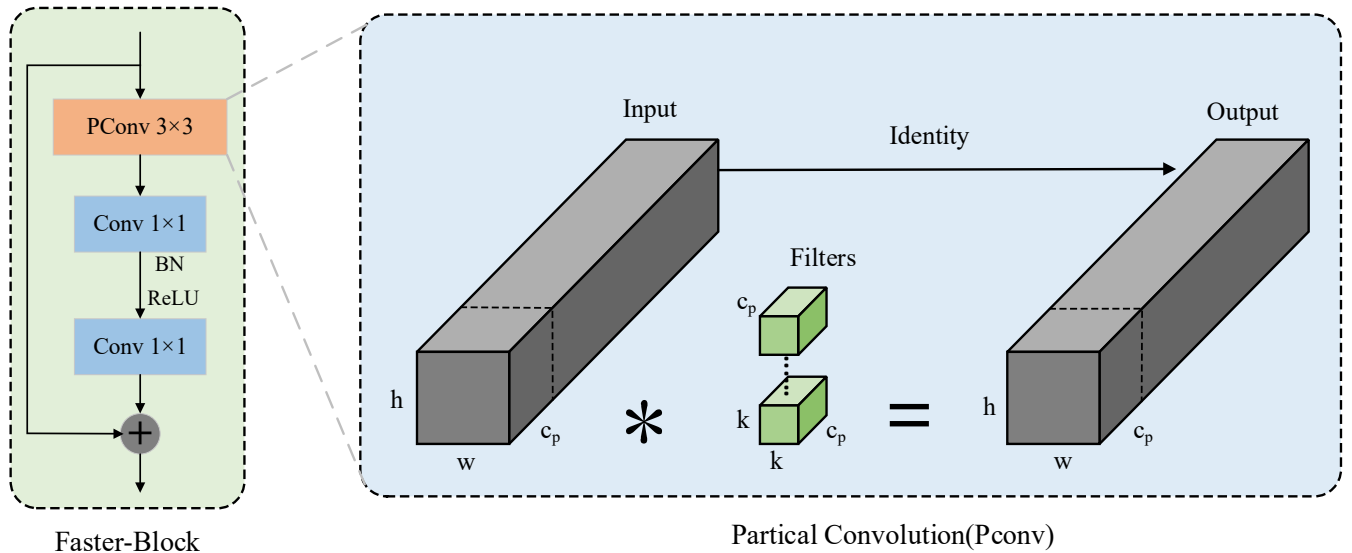

Fig. 15. Structure of the FasterNet Block module

To address the challenges posed by oversized models affecting detection speed and deployment difficulties on edge devices in flame detection tasks, the BottleNeck sections of the original YOLOv8s C2f module were replaced with the FasterNet Block module. This new module combines PConv and Conv layers, as shown in Figure 16. Compared to the original C2f module, the improved algorithm optimizes detection speed and computational complexity without compromising detection accuracy. The updated model reduces computational complexity and parameter count in object detection tasks, enabling flame detection on resource-limited devices such as development boards. The enhanced model provides rapid predictions for flame recognition while maintaining lightweight characteristics and adequate performance.

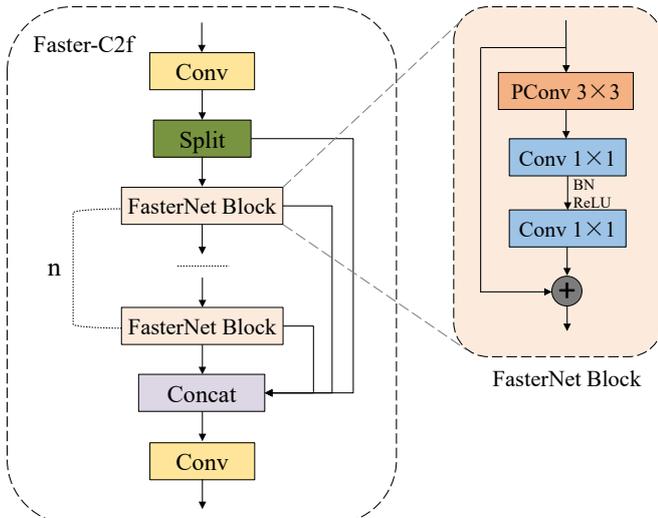

Fig. 16. Structure of the Faster-C2f module

The C2f module in the original YOLOv8s network was replaced with the Faster-C2f module, without altering the theoretical repetition count of the C2f module. This achieves a lightweight improvement of the YOLOv8s model. For clarity, the updated module is labeled as Faster-C2f, with the revised grid structure shown in Figure 17. To implement the addition and replacement of the FasterNet Block module, configuration changes are required in the block.py and tasks.py files within the YOLOv8 environment directory. Additionally, the "C2f" entries in the yolov8.yaml file must be updated to "Faster-C2f" to complete the lightweight network improvement and replacement. Faster-C2f integrates the FasterNet Block into the backbone network, aiming to enhance the model's detection speed.

## IV. EXPERIMENTAL COMPARISON AND ANALYSIS

### A. Experimental Environment

The experiments in this study were conducted on a Windows 10 Professional operating system, utilizing an NVIDIA GeForce RTX 3070 GPU with 8GB of VRAM. The CPU used was an Intel(R) Core(TM) i7-10700F @ 2.90GHz, paired with 16GB of RAM. Python 3.8.0 was employed as the programming language, with PyCharm serving as the development environment. The deep learning framework used for training was PyTorch 2.2.1 with cu121. The configuration of the training parameters is detailed in Table II.

TABLE II
EXPERIMENTAL PARAMETER SETTINGS

| Parameter Name | Parameter Settings |
| --- | --- |
| epochs | 150 |
| batch | 16 |
| workers | 8 |
| imgsz | 640 |
| optimizer | SGD |
| seed | 0 |
| lr0 | 0.01 |
| momentum | 0.937 |

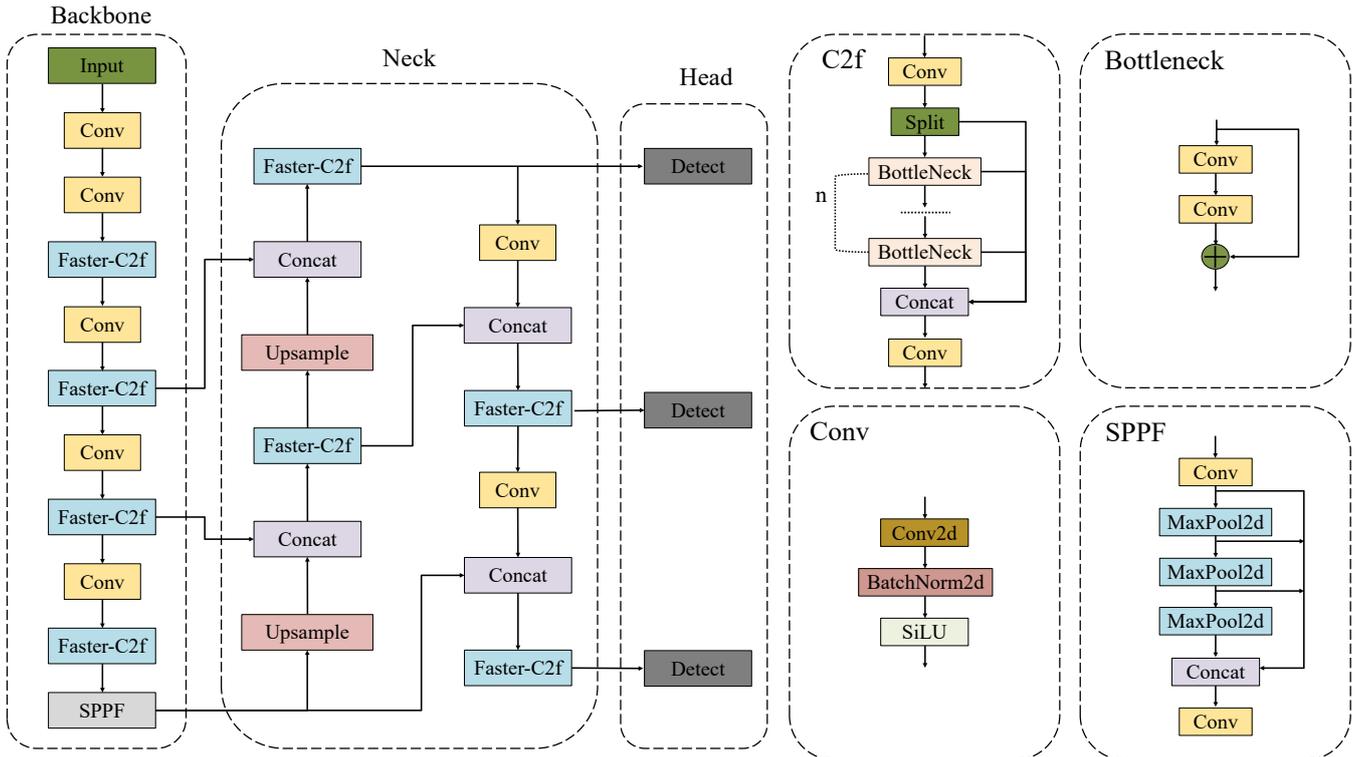

Fig. 17. Architecture of the Light-YOLOv8-Flame network

TABLE III
COMPARISON OF BENCHMARK ALGORITHMS

| Model | Precision | Recall | mAP@50 | FPS | Param/10⁶ | FLOPs |
|---|---|---|---|---|---|---|
| YOLOv8n | 73.61% | 60.16% | 66.27% | 82 | 3.00 | 8.1G |
| YOLOv8s | 76.22% | 61.35% | 67.39% | 77 | 11.13 | 28.4G |
| YOLOv8m | 76.72% | 60.79% | 66.03% | 66 | 25.84 | 78.7G |

### B. Baseline Algorithm Preliminary Experiments

To explore which version of the YOLOv8 series algorithms is better suited for flame detection scenarios, prior experiments were conducted under a unified framework using three versions of the YOLOv8 algorithm: YOLOv8n, YOLOv8s, and YOLOv8m, on a flame detection dataset. The comparison results are shown in Table III.

The experimental results indicate that the YOLOv8n model has the smallest computational load and the fastest detection speed, but its detection accuracy is relatively low. Although the YOLOv8m model shows some improvement in detection accuracy, it significantly increases computational load and reduces detection speed compared to the YOLOv8s model. YOLOv8s strikes a balance by maintaining high detection accuracy while achieving a high recall rate. Additionally, its detection speed is comparable to YOLOv8n, with moderate model size and computational resource requirements. Based on a comprehensive analysis, the YOLOv8s model is selected as the baseline model for this study.

### C. Loss Function Selection Experiment

YOLOv8's loss is mainly composed of classification and bounding box regression components. For classification loss, Cross Entropy (CE) is used, which aids the flame detection model in achieving more accurate classification during training. The calculation formula is given in Equation (6). For regression loss, Complete Intersection over Union (CIoU) [32] loss and Distribution Focal (DF) loss are employed. These components work together to optimize the model's performance in accurately localizing and classifying flame regions.

$$L_{CE} = -y\log\hat{y} - (1-y)\log(1-\hat{y}) \quad (6)$$

Calculating the bounding box regression loss is crucial to the performance of object detection. The model determines regression loss by computing the difference between the predicted bounding box and the ground truth box. In object detection, Intersection over Union (IoU) [33] is commonly used to measure the overlap between the predicted box (A) and the ground truth box (B). The calculation formula is provided in Equation (7).

$$IoU = \frac{|A \cap B|}{|A \cup B|} \quad (7)$$

However, traditional IoU loss functions have certain limitations. When the predicted box and the ground truth box do not overlap, the IoU is 0, resulting in a loss of 1, which does not account for the distance between the bounding boxes. Furthermore, even if two predicted boxes have the same IoU with the ground truth box, traditional IoU loss cannot distinguish which prediction is more accurate if their positions differ. To address these issues, YOLOv8 adopts the improved CIoU loss function for bounding box regression. CIoU loss not only considers IoU but also incorporates the distance between the center points of the bounding boxes. Additionally, it incorporates a penalty on center deviation and constrains aspect ratio variation, enabling more accurate and comprehensive assessment of predicted bounding box quality. The calculation formula for CIoU loss is given in Equation (8).

$$CIoU = IoU - \left(\frac{\rho^2(b, b^{gt})}{c^2} + \alpha v\right) \quad (8)$$

Here, $v$ and $\alpha$ are defined as shown in Equations (9) and (10). $v$ represents the aspect ratio consistency between the predicted and ground truth boxes, while $\alpha$ is a balancing coefficient. The value of $\alpha$ increases as the IoU between the predicted and ground truth boxes becomes larger.

$$v = \frac{4}{\pi^2}\left(\arctan\frac{w^{gt}}{h^{gt}} - \arctan\frac{w}{h}\right)^2 \quad (9)$$

$$\alpha = \frac{v}{(1-IoU)+v} \quad (10)$$

To determine which loss function is more effective for the flame detection task, a comparison of six loss functions, including CIoU, EIoU [34], and GIoU [35], was conducted based on the original YOLOv8s model. The results are shown in Figure 18.

The comparison results of the loss functions are presented in Table IV. The experimental results indicate that CIoU loss, used in the original YOLOv8s, achieves the highest precision and mAP, as well as superior average precision under different IoU thresholds compared to the other loss functions. This demonstrates the effectiveness of the CIoU loss function.

TABLE IV
EXPERIMENTAL COMPARISON OF LOSS FUNCTIONS

| Loss Function | Precision | Recall | mAP@50 | mAP@50-95 | F1-score |
|---|---|---|---|---|---|
| EIOU | 75.85% | 61.53% | 65.77% | 39.95% | 0.68 |
| SIOU | 74.89% | 61.06% | 66.03% | 39.97% | 0.67 |
| CIOU | 76.22% | 61.35% | 67.39% | 40.63% | 0.68 |
| GIOU | 74.54% | 62.61% | 66.58% | 40.51% | 0.68 |
| DIOU | 72.79% | 63.2% | 66.72% | 40.80% | 0.68 |
| WIOU | 72.66% | 62.41% | 65.47% | 39.55% | 0.67 |

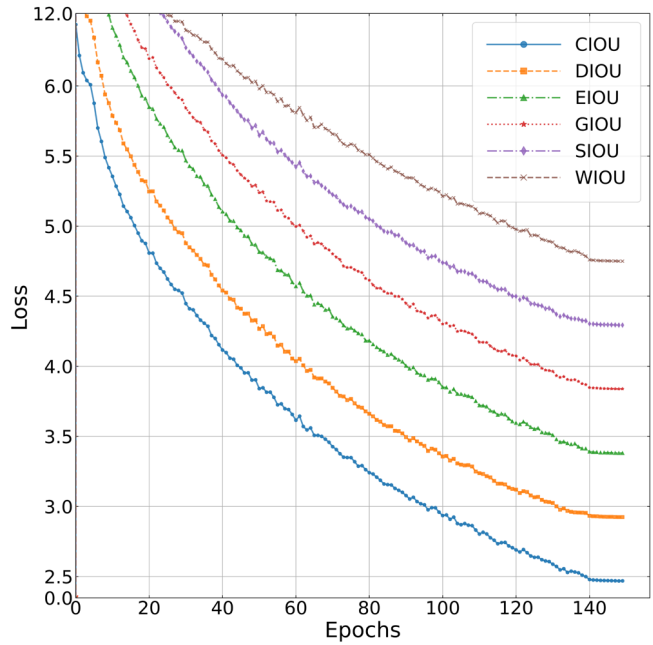

a) Loss curves

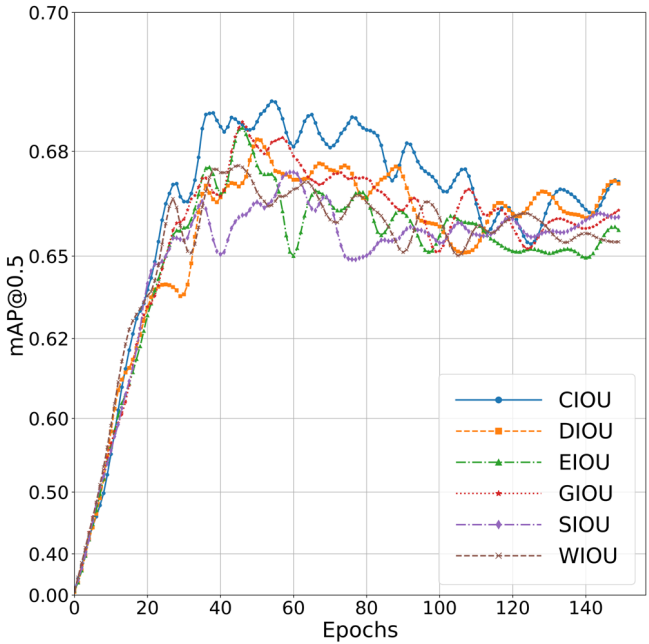

b) mAP@0.5 curves

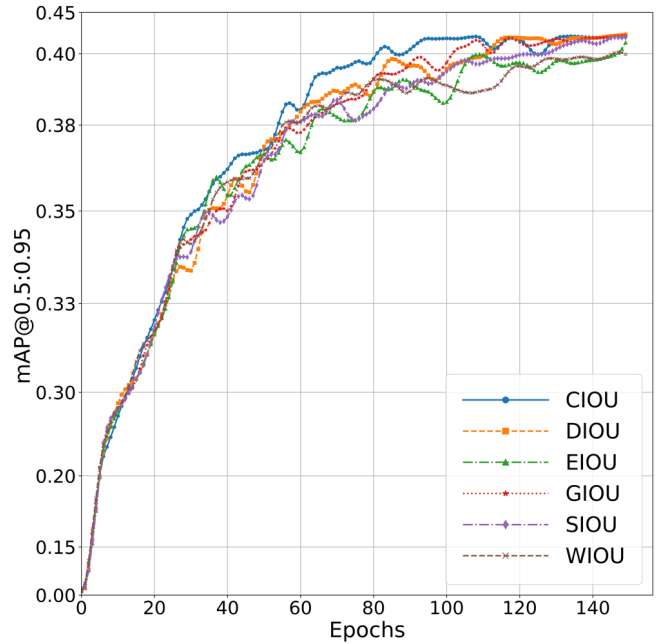

c) mAP@[0.5:0.95] curves

Fig. 18. Comparison of different loss functions

*D. Flame Detection Algorithm Comparison Experiment*

In this study, training was conducted on a self-constructed flame image dataset, and the proposed improved algorithm model was tested. Subsequently, it was compared with the original YOLOv8 model to validate the effectiveness of the new flame detection method.

*Ablation Experiment*

To evaluate the specific impact of each improved module on YOLOv8's performance optimization, ablation experiments were conducted. These experiments helped identify the contribution of each module to the overall performance. Table V presents the outcomes obtained from the ablation experiments.

TABLE V
ABLATION STUDY

| YOLOv8s | Faster-C2f | Precision | Recall | mAP@50 | FPS | Param/$10^6$ |
|---|---|---|---|---|---|---|
| √ | | 76.22% | 61.35% | 67.39% | 77 | 11.13 |
| √ | √ | 75.40% | 63.40% | 68.17% | 78 | 8.31 |

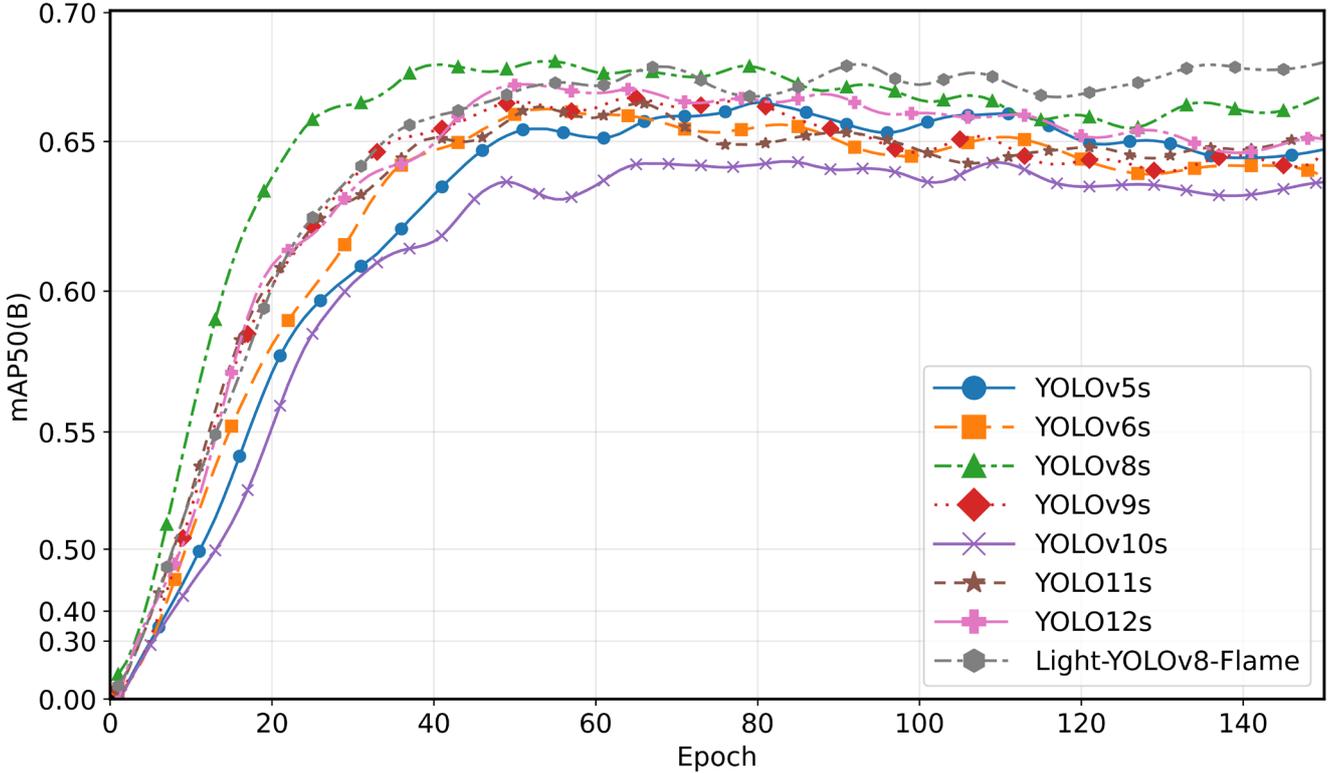

Fig. 19. mAP@50 performance of YOLO models over 150 epochs

By comparing various evaluation metrics, it was found that replacing the C2f module with the lightweight Faster-C2f module resulted in a 0.78% improvement in mAP and a 2.05% increase in recall rate, while reducing the parameter count by 25.34%. Precision decreased slightly by only 0.82%.

The experimental results demonstrate that replacing the C2f module with Faster-C2f achieves better performance in reducing parameter count and improving detection speed. Therefore, in flame detection tasks, if detection speed is a higher priority, the Light-YOLOv8-Flame model can be selected.

*Comparison Experiment of Mainstream Algorithms*

To thoroughly assess the effectiveness of the proposed algorithm, several comparative experiments have been carried out. The improved algorithm was compared with several current mainstream flame detection algorithms, including YOLOv5s, YOLOv6s, YOLOv8s, YOLOv9s, YOLOv10s, YOLOv11s, and YOLO12s. The mAP@50 of each model as it changes with respect to epochs is shown in Figure 19. Additionally, Table VI displays the outcomes of the comparison.

The experimental data indicate that while YOLOv5s is relatively lightweight in terms of parameter count, its performance across evaluation metrics is moderate. YOLOv6s demonstrates an advantage in detection speed but comes with relatively higher computational complexity. YOLOv9s offers a balanced solution with reasonable performance metrics but underperforms in mAP@50 when compared to other models. YOLOv10s achieves a reasonable trade-off between accuracy and speed, although it falls short in both dimensions. YOLOv11s performs well in both speed and accuracy, but its computational complexity is higher than that of YOLOv5s or YOLOv10s. YOLOv12s is similar to YOLOv11s, offering a balanced trade-off, though with slightly lower precision and recall. In contrast, the proposed improved algorithm, while slightly less accurate than the original YOLOv8s and with a slight reduction in detection speed, outperforms the other flame detection models in terms of mAP, parameter count, and computational complexity. From a comprehensive perspective, considering mAP, computational complexity, and parameter count, the proposed improved algorithm demonstrates superior performance compared to other mainstream algorithms.

TABLE VI
PERFORMANCE COMPARISON OF MAINSTREAM ALGORITHMS

| Model | Precision | Recall | mAP@50 | FPS | Param/$10^6$ | FLOPs |
| --- | --- | --- | --- | --- | --- | --- |
| YOLOv5s | 74.81% | 61.04% | 64.93% | 70 | 9.11 | 23.8G |
| YOLOv6s | 74.37% | 61.53% | 63.88% | 83 | 16.30 | 44.0G |
| YOLOv8s | 76.22% | 61.35% | 67.39% | 77 | 11.13 | 28.4G |
| YOLOv9s | 72.61% | 63.59% | 64.69% | 80 | 7.17 | 26.7G |
| YOLOv10s | 72.97% | 61.83% | 63.81% | 75 | 8.04 | 24.4G |
| YOLOv11s | 74.45% | 62.12% | 65.46% | 82 | 9.41 | 21.3G |
| YOLOv12s | 72.67% | 63.40% | 65.25% | 79 | 9.23 | 21.2G |
| Light-YOLOv8-Flame | 75.40% | 63.40% | 68.17% | 78 | 8.31 | 21.4G |

## V. CONCLUSION

This paper presents Light-YOLOv8-Flame, an optimized method based on the YOLOv8 algorithm, designed to enhance fire detection accuracy and response speed. Firstly, a comprehensive flame image dataset was constructed, encompassing multiple scenes, to effectively address the limitations of existing flame detection models in complex environments. Secondly, the model's structure was optimized by replacing the FasterNet Block residual connection in YOLOv8, thereby enhancing its flame detection capabilities. This modification not only improved the model's efficiency but also provided a more detailed analysis of image features. Experimental results demonstrate that the optimized algorithm performs exceptionally well in fire detection tasks, achieving a 0.78% increase in mAP while reducing the model's parameter count by 25.34%. These improvements illustrate that the model enhances detection performance while maintaining a lightweight design. This research advances both the scientific and practical aspects of fire detection technology, offering an innovative solution for fire monitoring. The findings contribute to the improvement of flame detection algorithms, with significant application potential and societal implications.